\documentclass{article}
\usepackage{spconf,amsmath,graphicx}
\usepackage{amsmath}
\usepackage{amssymb}

\usepackage{mathtools}
\usepackage{multirow}

\usepackage{amsmath,amssymb,amsfonts,mathtools}
\usepackage{textcomp,graphicx,multirow,cite}
\usepackage{algpseudocode}

\graphicspath{{./images/}}

\newcommand{\pp}{\mathbf{p}}
\newcommand{\qq}{\mathbf{q}}
\newcommand{\uu}{\mathbf{u}}
\newcommand{\Aset}{\mathcal{A}}
\newcommand{\Bset}{\mathcal{B}}

\usepackage{algorithm}

\usepackage{subcaption}
\title{On Segmentation of Pectoral Muscle in Digital Mammograms by means of Deep Learning}
\name{Hossein Soleimani, Oleg V. Michailovich}
\address{Department of Electrical and Computer Engineering, University of Waterloo\\ (e-mail: h3soleim, olegm@uwaterloo.ca)}
\begin{document}
%
\maketitle
\begin{abstract}
Computer-aided diagnosis (CAD) has long become an integral part of radiological management of breast disease, facilitating a number of important clinical applications, including quantitative assessment of breast density and early detection of malignancies based on X-ray mammography. Common to such applications is the need to automatically discriminate between breast tissue and adjacent anatomy, with the latter being predominantly represented by pectoralis major (or pectoral muscle). Especially in the case of mammograms acquired in the mediolateral oblique (MLO) view, the muscle is easily confusable with some elements of breast anatomy due to their morphological and photometric similarity. As a result, the problem of automatic detection and segmentation of pectoral muscle in MLO mammograms remains a challenging task, innovative approaches to which are still required and constantly searched for. To address this problem, the present paper introduces a two-step segmentation strategy based on a combined use of data-driven prediction (deep learning) and graph-based image processing. In particular, the proposed method employs a convolutional neural network (CNN) which is designed to predict the location of breast-pectoral boundary at different levels of spatial resolution. Subsequently, the predictions are used by the second stage of the algorithm, in which the desired boundary is recovered as a solution to the shortest path problem on a specially designed graph. The proposed algorithm has been tested on three different datasets (i.e., MIAS, CBIS-DDSm and InBreast) using a range of quantitative metrics. The results of comparative analysis show considerable improvement over state-of-the-art, while offering the possibility of model-free and fully automatic processing.
\end{abstract}

\section{Introduction}\label{sec:introduction}
{B}{reast} cancer (BC) is the most widespread malignancy in women worldwide. Even though since the introduction of screening X-ray mammography the mortality from BC has been reduced by more than 40$\%$, the disease keeps claiming around 400,000 lives around the globe every year \cite{ferlay2010global}. To further reduce the fatality rates requires application of more accurate methods of detection of breast disease in screening mammograms. Unfortunately, due to the relatively low contrast of the imaging modality, such detection is known to be a difficult task, especially in the case of (radiographically) dense breast \cite{sprague2014prevalence}. Moreover, analysis and interpretation of large amounts of mammographic data may be a challenging task for the radiologists, in which case they often opt to rely on {\it computer-aided diagnosis} (CAD) systems which render the problem far more manageable. Nowadays, CAD systems play an increasingly important role in detection and classification of breast lesions, especially in their early pathological stages, where they might be missed or overlooked by a human interpreter  \cite{hadjiiski2006advances}.

\begin{figure*}[t]
\centerline{\includegraphics[width=17.8cm]{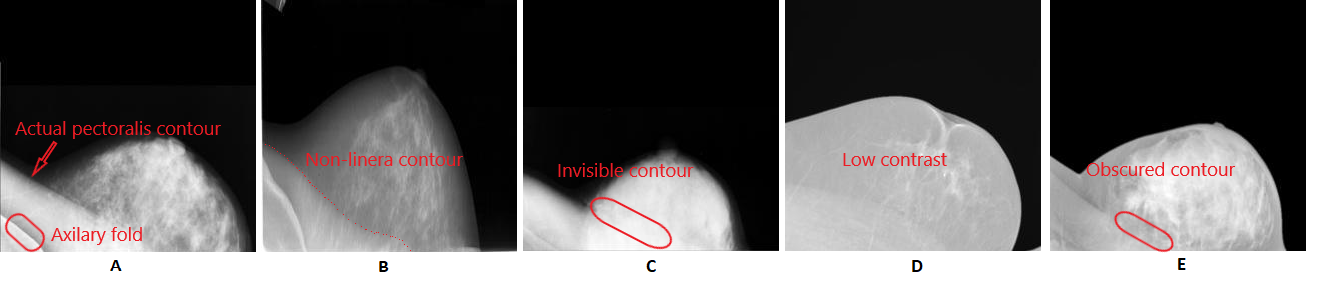}}
\caption{Varying appearance of the pectoral boundary due to the effects of inter-subject variability.}
\label{Challenges}
\end{figure*}

At present, quantitative assessment of breast density as well as cross-modal registration of mammographic scans with the images produced by other modalities (such as, e.g., ultrasound or magnetic resonance imaging) are among the standard applications of CAD in radiological management of BC \cite{shi2018hierarchical, rampun2019segmentation, soleimani2019segmentation}. At their initial stages, all such tools involve the process of identifying the image domain associated with the breast tissue -- the process commonly referred to as {\it breast segmentation}. In particular, in mammograms acquired in the medial-lateral oblique (MLO) view, the posterior boundary of breast tissue interfaces with pectoralis major which is often used as a key landmark for establishing the breast-body bounding line. Unfortunately, the delineation of pectoralis major in MLO mammograms is known to be a difficult problem for a number of reasons. In particular, both in terms of its morphology and contrast, the pectoral muscle can be similar to the appearance of fibroglandular tissue, which renders their delineation quite problematic. This situation is particularly frequent in the case of dense breast tissue whose fibroglandular component tends to edge near the breast-body bounding line, often overlapping (and, as a result, obscuring) the pectoral region. Moreover, due to significant inter-subject variability, the boundary of pectoralis major does not have a consistent appearance in MLO scans, exhibiting substantial variations in shape (i.e., from quasi-linear to curved) and/or visibility due to low imaging contrast and physical occlusions (see Fig.~\ref{Challenges} for the examples of clinical mammograms which demonstrate the above effects). Consequently, the application of conventional methods of statistical shape analysis have shown rather limited ability to improve the accuracy of breast segmentation.

\section{Related works}
As mentioned above, the main in challenge in the problem of breast segmentation in MLO mammograms stems from the non-triviality of detection of pectoral boundary. It is therefore not a surprise that the detection problem has been addressed in multiple studies using a range of different approaches. In particular, there have been a group of methods based on specific {\it a priori} assumptions regarding the boundary geometry. Thus, for example, in \cite{karssemeijer1998automated}, the boundary was approximated by a straight line, thus allowing the authors to take advantage of the Hough transform for its detection. While very efficient numerically, however, this approach suffers from the simplicity of its model assumptions, which limits its applicability in the case of curved boundaries (e.g., as shown in Subplot B of Fig.~\ref{Challenges}). 

The limitations of the straight-line model of \cite{karssemeijer1998automated} have been addressed by a number of later studies. Several of these studies used the straight-line model as an initialization which was subsequently 
refined into a curved configuration by means of additional processing steps. In particular, the refinement based on the use of active contours was described in \cite{ferrari2004identification} and \cite{kwok2004automatic}, while in \cite{bora2016robust} the authors took advantage of polynomial fitting within a regression framework. The above approaches, however, depend on the accuracy of the initialization which is known to be sensitive to the morphological properties of pectoral muscle.

The initialization of pectoral boundary based on the results of adaptive thresholding has been discussed by several authors as well \cite{mustra2013robust, chen2010segmentation, vikhe2017detection, yin2019robust}. In particular, following the thresholding stage, the initial pectoral boundary was refined through curve fitting using least squares minimization in \cite{vikhe2017detection}. A more advanced approach was proposed in \cite{yin2019robust}, where the segmentation stage was preceded by image enhancement by means of fractional differentiation, followed by iterative thresholding, initial curve fitting and subsequent active contour evolution. In \cite{chen2015shape}, the segmentation was obtained by means of a region-growing algorithm in combination with a cubic polynomial model, while in \cite{yoon2016automatic}, the authors took advantage of morphological image processing and the random sample consensus (RANSAC) algorithm. However, the dependence of the above methods on photometric information and low-order polynomial modelling makes them overly sensitive to the appearance (visibility) of the pectoral muscle. As a result, their performance may still deteriorate dramatically when the pectoral boundary is either visually indiscernible or obscured by dense tissue, which is a common effect of tissue folding.

The advent of {\it deep learning} (DL) and, in particular, of {\it convolutional neural networks} (CNN), has offered conceptually new possibilities to tackle the problem of breast segmentation. Subsequently, several studies aimed at designing the CNN architectures which were specifically tailored to the problem at hand. Among such works, the most promising results have been demonstrated in \cite{dubrovina2018computational, moeskops2016deep, rampun2019breast}. More specifically, in \cite{dubrovina2018computational, moeskops2016deep}, the CNNs were trained using image blocks (i.e., mini patches) extracted from the pectoral region, thus taking into account the structural appearance and photometric properties of the pectoral muscle. Despite the promising results reported by these two studies, their output segmentation maps still suffered from the effects of false positive misclassification. To overcome this problem, it was proposed in \cite{rampun2019breast} to subject the output of CNN-based classification to a post-processing stage involving morphological operations. It is worthwhile noting that, in its first stage, the method relied on a modified version of the hierarchical edge detection (HED) network of \cite{xie2015holistically} which is capable of integrating the information on the location of pectoral edges across multiple resolution scales. 

It should be pointed out that, the last few years have seen a rapid development of edge detection CNNs which are capable of detecting both fine- and course-scale representations of various geometric structures in data images (such as, e.g., edges, ridges, etc) \cite{liu2019richer, poma2020dense}. The main idea of these methods is to produce edge maps in different resolution scales, followed by fusing the obtained information to yield the final result. However, while very effective for reducing uncertainties (due to imaging artifacts and the effects of noise), the process of fusion is not without drawbacks, chief of which is often attributed to the excessive thickness of resulting edges \cite{poma2020dense}. This could be a serious disadvantage in the case of breast segmentation, in which case the edges are required to be as fine as possible. 

Motivated by the results of \cite{rampun2019breast}, the present study proposes a new approach to the problem of breast segmentation which, similarly to earlier works in the field, relies on a two-stage processing scheme. The first stage uses a modified version of the VGG16 network architecture \cite{simonyan2014very}. The introduced modification restricts the analysis to two resolution levels and, as a result, it has the important advantage of depending on a relatively small number of network parameters which are substantially easier to train. Moreover, the proposed architecture proves to be sufficient to encode all the relevant information about the pectoral boundary, with the course level pinpointing its anatomical location (with virtually no ``false positives") and the fine level following its true configuration at a much higher resolution. Although the fine resolution map remans prone to numerous misclassification errors, the latter are effectively eliminated by the fusion process, yielding an accurate initialization for the second stage of the proposed algorithm. At this stage, we construct a weighted graph and take advantage of Dijkstra's algorithm to locate the pectoralis boundary at a {\it single-pixel} resolution (thickness). The proposed algorithm is fully automatic and capable of reliably detecting the breast-body interface in complex scenarios (such as shown in Subplots C and D of Fig.~\ref{Challenges}), when alternative methods tend to fail.
The proposed solution has been tested on three public datasets, containing both full filled digital mammograms and scanned film mammograms, with the results confirming the effectiveness of the new approach.

The remainder of the paper is organized as follows. The proposed method, including description of the designed CNN architecture and post-processing, is summarized in Section \ref{sec_proposed_method}, while section \ref{sec_results} presents the quantitative and qualitative results of our experimental study. Section \ref{sec_conclusion} concludes the paper with a discussion of its principal findings.

\section{Proposed Method} \label{sec_proposed_method}
The proposed approach consists of two principal steps: (a) CNN-based delineation of the breast-body interface (i.e., pectoral boundary) and estimation refinement by means of a graph search algorithm. This section will discuss these steps in details.

\subsection{CNN-based edge detection}
The CNN used in the present study follows the design principal of the neural networks which have been recently reported in \cite{xie2015holistically, liu2019richer, poma2020dense}. In particular, all these methods take advantage of multiple CNNs which correspond to different resolution levels and are optimized using different number of training images. The backbone of these models is formed by the convolutional blocks of VGG16 architecture, where edge maps are generated from each block resulting in a multi-scale learning structure. Thus, for example, the HED network of \cite{xie2015holistically} exploits a total of five convolutional blocks, taking a side-output from each of them. Subsequently, the outputs are integrated by a fusing block producing the final edge map. Note that each of the convolutional block as well as the fusing block are optimized using different cost functions (i.e., a total of six in the case of \cite{xie2015holistically}), which requires a relatively large number of training samples.

\begin{figure}[t]
  \centering
  \centerline{\includegraphics[width=8.5cm]{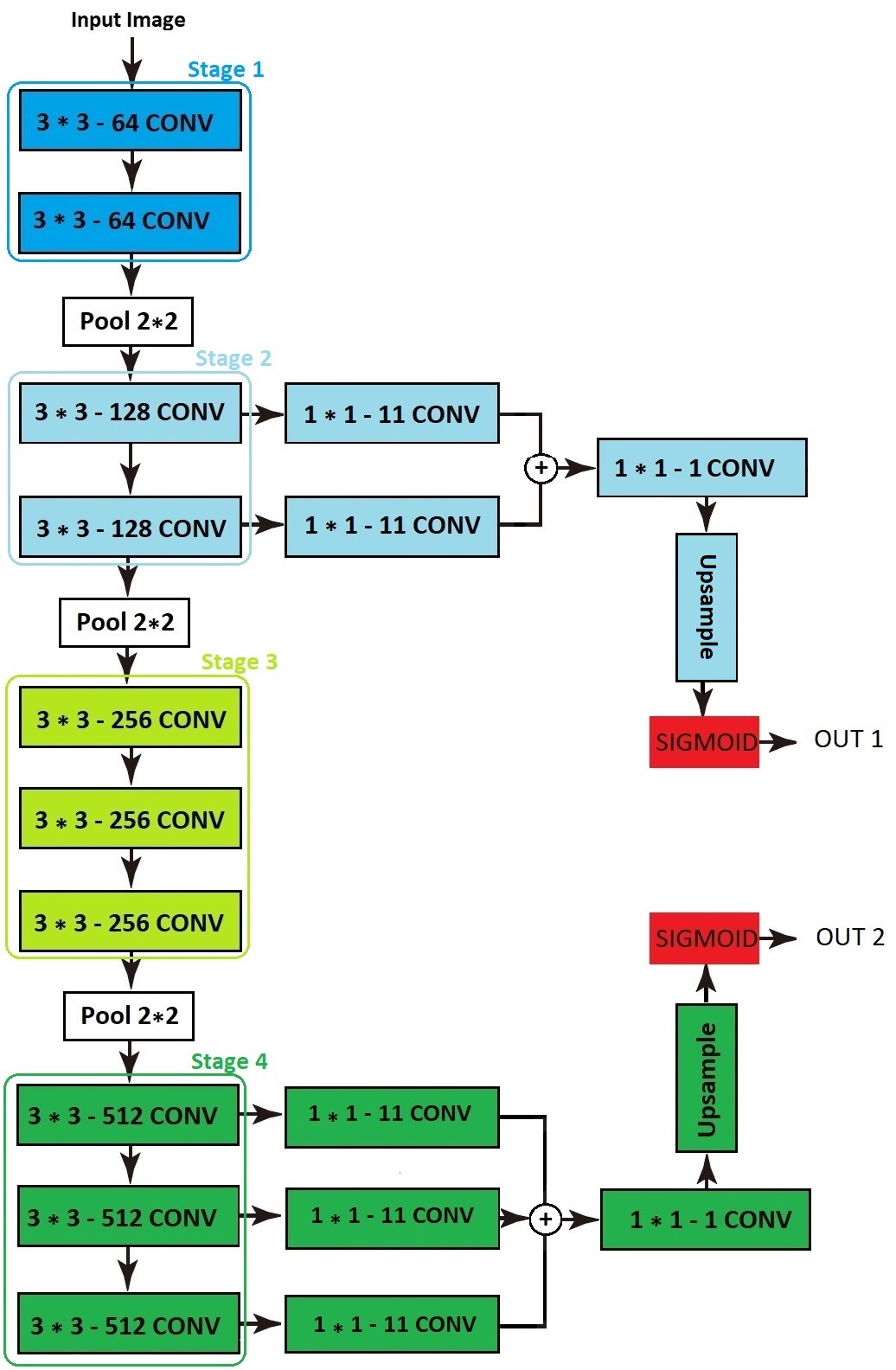}}
\caption{Proposed CNN architecture for pectoral boundary detection.}
\label{fig_archit}
 \end{figure}

In the above-mentioned HED network, the edge (probability) maps are computed at different resolution levels, resulting in edge estimates of variable thickness. In this case, the final estimate produced by the process of fusion may not always reduce the edge thickness to a single-pixel resolution \cite{poma2020dense}. However, this seems to be a reasonable compromise, considering the superb performance of the network in complex scenarios involving natural images. In the present work, however, the network is only required to produce a sufficiently accurate initialization, which can be achieved using a simplified network architecture (with a substantially reduced number of training parameters), as explained next. 

The proposed network uses VGG16 as its base element. The latter consists of three fully-connected (FC) layers and 13 convolutional ({\it conv}) layers which are subdivided into five stages, each of which is followed by a pooling layer. The proposed architecture, on the other hand, excludes the 5th stage along with all the FC layers, resulting in a simplified design depicted in Fig.~\ref{fig_archit}.

To further reduce the network complexity, the proposed network is designed to output the edge maps pertaining to stages 2 and 4 which are left unfused. Instead, the convolutional layers of stages 2 and 4 are connected to a {\it conv} layer associated with a $1\times 1$ kernel of depth 11. Note that, in the original VGG16 scheme, the kernel depth is set to 21, which leads to an unnecessary higher complexity for the problem at hand.  

The feature maps produced at each stage are then passed to another {\it conv} layer with an associated kernel of size $1\times1$ and depth 1. Finally, the resulting feature maps are resampled to the original resolution of input images (using linear interpolation) and fed to a {\it softmax} function to produce two edge probability maps. The latter are used by the second step of the proposed method to locate the pectoral muscle.

\begin{figure*}[t]
  \centering
  \centerline{\includegraphics[width=18cm]{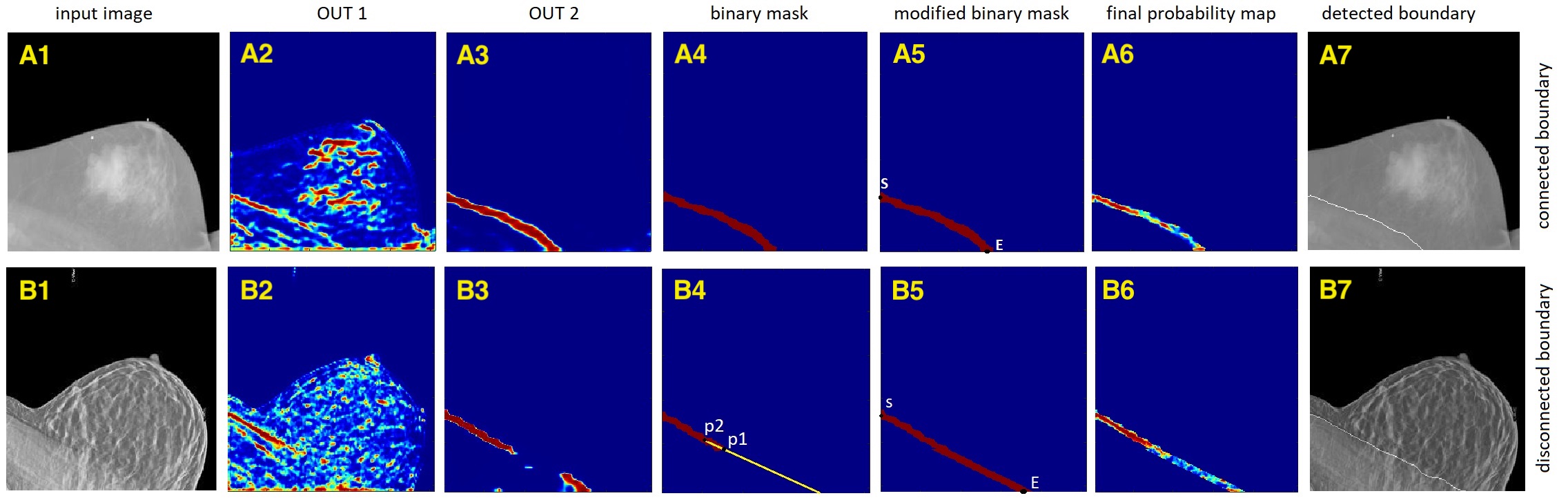}}
\caption{Upper row of subplots: (A1) input MLO mammogram, (A2) edge probability map OUT 1, (A3) edge probability map OUT 2, (A4) binary mask $B$, (A5) modified binary mask, (A6) final edge probability map, and (A7) the result of graph-based edge detection. Subplots B1-B7 are composed in an analogous manner, corresponding to a different input image shown in Subplot B1.}
\label{fig_results}
 \end{figure*}

\subsubsection{Lost function}
Let $W$ denote the array of network parameters to be optimized. The objective function used for training the proposed CNN can then be expressed as
\begin{equation}
    \mathfrak{L}(W)=\mathfrak{L}_1(X^1,W) + \mathfrak{L}_2(X^2,W)
\end{equation}
where $X^1$ and $X^2$ are the activation values (feature maps) produced by stages 2 and 4 of the network. In this formulation, $\mathfrak{L}_1(X^1,W) $ and $ \mathfrak{L}_2(X^2,W)$ represent the lost functions of the two stages, respectively. Each lost function is computed over all pixels of a training image and its reference edge map $Y$. The latter has the form of a binary mask which assumes the value of one at the location of the true boundary, while being equal to zero for other pixels, i.e., $Y_i \in \{0,1\},\:\: i=1,2, \ldots, |Y|$. For obvious reasons, the majority of these labels always correspond to non-contour pixels, which introduces a considerable bias towards the ``out-of-boundary" pixels. To compensate for this undesirable effect, the loss for each pixel $X_i$ in activation $X$ with respect to its label $Y_i$ is computed as
\begin{equation}
 l(x_i) = 
  \begin{cases}
   \alpha \cdot \log(1-\mbox{sgm}(x_i)) & \text{if } y_i=0 \\
   \beta \cdot \log(\mbox{sgm}(x_i)) & \text{if } y_i=1
  \end{cases}
\end{equation}
where $\rm sgm$ stands for the standard sigmoid function, and 
\begin{equation}\label{balance}
\alpha = \lambda \frac{|Y^+|}{|Y^+| +|Y^-|} \:\:,\:\:
\beta =  \frac{|Y^-|}{|Y^+| +|Y^-|},
\end{equation}
with $\lambda$ being a positive tuning parameter used to balance the discrepancy in the number of ``out-of-boundary" pixels ($|Y^-|$) and ``on-boundary" pixels ($|Y^+|$).

\subsubsection{Network training}
The proposed network was trained and tested using three public datasets, namely MIAS \cite{suckling1994mammographic}, InBreast \cite{moreira2012inbreast}, and CBIS-DDSM \cite{lee2017curated}. MIAS and CBIS-DDSM consist of 322 and 457 scanned film mammograms, respectively, of different image sizes. InBreast dataset, on the other hand, consists of 208 full-filled digital mammograms (FFDM) of size $2560\times3328$. In order to create the reference edge labels (i.e., $Y$), all the mammograms have been manually annotated by a qualified radiologist. Both data mammograms and their related edge maps were subsequently resized to a standard dimension of $256\times256$ pixels. 

The availability of three different datasets has allowed us to employ a ``2 + 1" validation strategy, in which two datasets (e.g., InBreast and CBIS-DDSM) were used fo the purpose of network training, followed by predicting the labels of the the third ``unseen" dataset (e.g., MIAS). Note that, in addition to fair assessment of validation errors, the above strategy has an important advantage of providing a useful insight into the effects of between-scanner variability. 

To minimize the effects of overfitting, the experimental dataset was extended by means of data augmentation, which is a standard practice in DL \cite{shorten2019survey}. To this end, each training image was resized by factors 0.9 and 1.1, followed by either cropping or zero-padding of the results thus obtained to the target size of $256\times256$ pixels. Additional training images were obtained from the original mammograms through the process of vertical, horizontal, and diagonal flipping. In this way, the size of the training set was increased by a factor of 10.   

To initiate training, we used the VGG16 weights optimized over the ImageNet dataset \cite{imagenet_cvpr09}. The learning rate and the batch size were set to $0.0001$ and  $2$, respectively. The training was performed by means of stochastic gradient descent optimization \cite{ruder2016overview}, with the number of epochs set to 30.

The edge probability maps produced by the neural network will be denoted below by OUT 1 and OUT 2 (corresponding to stage 2 and 4, respectively). Two examples of such maps are shown in Subplots A2-A3 (respectively,  B2-B3) of Fig.~\ref{fig_results} for the input MLO mammogram shown in Subplot A1 (respectively, B1) of the same figure. Note that the probability maps are depicted in pseudo-colour, with the values of 0 and 1 represented by the blue and red colours, respectively.  

\subsection{Detection of pectoral boundary} \label{sec_dijkstra}
As shown in Fig.~\ref{fig_results}, the edge probability map OUT 1 provides a finer localization of the pectoral boundary as compared to OUT 2. At the same time, OUT 1 suffers from a substantially larger amount of clutter noise due to the network's response to the dense structures of breast tissue. In order to suppress the noise-related artifacts, therefore, the two maps need to be properly fused. To this end, we first binarize OUT 2 by means of hard (uniform) thresholding, resulting in a binary mask $B$. Subsequently, $B$ is subjected to morphological pruning that retains the longest connected component of $B$, which is normally associated with the pectoral boundary. Note that, as opposed to the majority of earlier approaches, the pruning step does not rely on any {\it a priori} assumptions regarding the orientation of pectoral muscle.

For numerical convenience, all the images were reoriented so as to position the pectoral region at the low-left corner of the image coordinate system (as shown in Fig.~\ref{fig_results}). As a result, the opposite ends of pectoral boundary had been constrained to lie on the first column and the last row of $B$, respectively (as shown in Subplots A4 and B4 of Fig.~\ref{fig_results}). 

Unfortunately, the numerical procedure described above has one critical limitation. In particular, it has been assumed that the longest connected component of $B$ coincides with the pectoral boundary {\it along its entire length} (as illustrated by Subplot A4 in Fig.~\ref{fig_results}). Unfortunately, situations are possible in which this component can be disconnected (as exemplified in Subplot B4 of the same figure).

The above problem can be overcome by means of simple linear extrapolation. In particular, suppose the edge indicated by $B$ is disconnected on the right and, as a result, its terminating point $p_1$ falls short of reaching the boundary of the image domain, as depicted in Subplot B4 of Fig.~\ref{fig_results}. The point $p_1$ can be defined by the (right) endpoint of the morphological skeleton of the partial edge. Furthermore, following the same skeleton (starting at $p_1$), one can define another point $p_2$, lying at a predefined arc-length distance $\mathcal{D}$. In the course of our experimental study, the value of $\mathcal{D} = 25$ was found to provide stable and consistent results.

Finally, the skeletal points within the interval defined by $p_1$ and $p_2$ can be fit by a line segment, which can, in turn, be used to complete the partial edge (see the illustration in Fig.~\ref{fig_results}). In the present study, the fitting was based on a standard LS formulation. Also, the thickness of the completing part of the edge was set to be equal to the average thickness of its initial segment. The result of edge completion is shown in Subplot B5 of Fig.~\ref{fig_results}.

Needless to add, the above considerations pertain to only one possible scenario, while in practice, the edge incompletion problem may arise on either or both sides of the longest connected component of $B$. In such cases, the edge completion can be done in an analogous way {\it mutatis mutandis}.  

In the end, given the binary mask $B$ (which could have been modified through the above described edge completion procedure, if necessary), the final edge probability map $M$ can be defined to be the result of element-wise product of $B$ with OUT 2. Formally,
\[
M = B \odot \mbox{OUT 2},
\] 
where $\odot$ denotes the element-wise (Hadamard) product between two equally-sized matrices. 

The computation of $M$ concludes the first phase of the proposed method. At the next stage, $M$ is used to resolve the breast-body interface at a single-pixel resolution. To this end, $M$ can first be converted to a symmetric, weighted, fully-connected graph $\mathcal{G}(V, E)$, with $V$ and $E$ denoting the nodes and edges of the graph, respectively.

The connectivity structure of $\mathcal{G}$ can be alternatively defined in terms of its associated matrix of (edge) weights $W$, which quantifies a degree of affinity between any two nodes of the graph. In this case, lower values of $W$ indicate ``stronger" connections, with $W({\bf p}, {\bf q}) = \infty$ representing the situation when nodes $\bf p$ and $\bf q$ are considered to be disconnected.

To facilitate the definition of $W$, we associate the nodes $V$ of $\mathcal{G}$ with {\it all} the pixels within the image domain, in which case $V$ becomes a uniform rectangular lattice. For each ${\bf p} \in V$, let $\mathcal{N}({\bf p})$ denotes its 8-connected neighbourhood and, with a slight abuse of notations, let the modified binary mask $B$ embody the subset of $V$ over which its values are equal to 1 (see Subplots A5 and B5 of Fig.~\ref{fig_results}). Then, for each $\pp \in V$, the $({\bf p},{\bf q})$-th element $W$ can be defined as
\begin{equation}\label{eq_graph}
W(\pp,\qq)=\begin{cases}
\frac{2}{M(\pp)+M(\qq)}, \quad &\mbox{if} \,\, \qq \in \mathcal{N}(\pp) \cap B \\
\infty, \quad &\mbox{otherwise}
\end{cases},
\end{equation}
where $\cap$ stands for set intersection.

\begin{figure}[t]
\caption{Modified Dijkstra's algorithm with backtracking}
\begin{algorithmic}[1]
\Procedure{Dijkstra}{$V, B, S, E$}
\State ${\rm dist}(S) \gets 0$
\State ${\rm dist}(\pp) \gets \infty, \forall \pp \in V \backslash \{S\}$
\State ${\rm parent}(\pp) \gets {\tt NIL}, \forall \pp \in V$
\State $\Aset \gets \{S\}$
\State $\uu \gets S$
\While{$\uu \neq E$}
\State $\Bset = \emptyset$
\For{$\pp \in \Aset$}
\For{$\qq \in (\mathcal{N}\{\pp\} \cap B)  \backslash \Aset$}
\If{${\rm dist}(\qq) > {\rm dist}(\pp) + W(\qq,\pp)$} 
\State ${\rm dist}(\qq) \gets {\rm dist}(\pp) + W(\qq,\pp)$
\State $\Bset \gets \Bset \cup \{\qq\}$
\EndIf
\EndFor 
\EndFor 
\State $\uu = \arg\min_{\qq \in \Bset} {\rm dist}(\qq)$
\State $\Aset \gets \Aset \cup \{\uu\}$
\For{$\qq \in (\mathcal{N}\{\uu\} \cap B)$}
\If{${\rm dist}(\qq) > {\rm dist}(\uu) + W(\qq,\uu)$} 
\State ${\rm dist}(\qq) \gets {\rm dist}(\uu) + W(\qq,\uu)$
\State ${\rm parent}(\qq) \gets \uu$
\EndIf
\EndFor
\EndWhile
\State $\mathcal{P} \gets \{E\}$
\State $\pp \gets E$
\While{$\pp \neq S$}
\State $\pp = {\rm parent}(\pp)$
\State $\mathcal{P} \gets \mathcal{P} \cap \{\pp\}$
\EndWhile \\
\Return $\mathcal{P}$
\EndProcedure
\end{algorithmic}
\end{figure}

\begin{figure}[t]
\centering
\caption{Pectoral boundary reconstruction}
\begin{algorithmic}[1]
\Procedure{Boundary}{OUT1, OUT2}
\State Compute $B$ from OUT2 via thresholding
\If{$B$ is disconnected} \\
\hspace{1cm}Complete $B$ via linear extrapolation
\EndIf
\State Compute $M = B \odot \mbox{OUT 2}$
\State Construct $\mathcal{G}$; define $S$ and $E$.
\State Compute $\mathcal{P}$ = {\sc Dijkstra}$(V,B,S,E)$ \\
\Return $\mathcal{P}$
\EndProcedure
\end{algorithmic}
\end{figure}

The above definition of edge weights is based on the values of edge probability map $M$. It is worthwhile noting that $M$ ``inherits" the best characteristics of both OUT 1 and OUT 2. In particular, similarly to OUT 1, the non-trivial values of $M$ are localized in close proximity of the true pectoral boundary. At the same time, similarly to OUT 2, $M$ remains immune to the effects of clutter noise. Thus, for any $\pp \in B$ and $\qq \in \mathcal{N}(\pp) \cap B$, the value of $W(\pp,\qq)$ is bound to decrease {\it pro rata} with an increase in the empirical probability of nodes $\pp$ and $\qq$ to lie near the true pectoral boundary. It is, therefore, not unreasonable to assume the latter to be associated with an {\it open path} on $\mathcal{G}$ formed by its most ``connected" nodes. More specifically, given two points $S$ (for ``start") and $E$ (for ``end") on the opposite sides of pectoralis, its boundary can be closely approximated by the {\it shortest path} on $\mathcal{G}$ that connects $S$ and $E$.  

\begin{figure*}[t]
\centering
\centerline{\includegraphics[width=18cm]{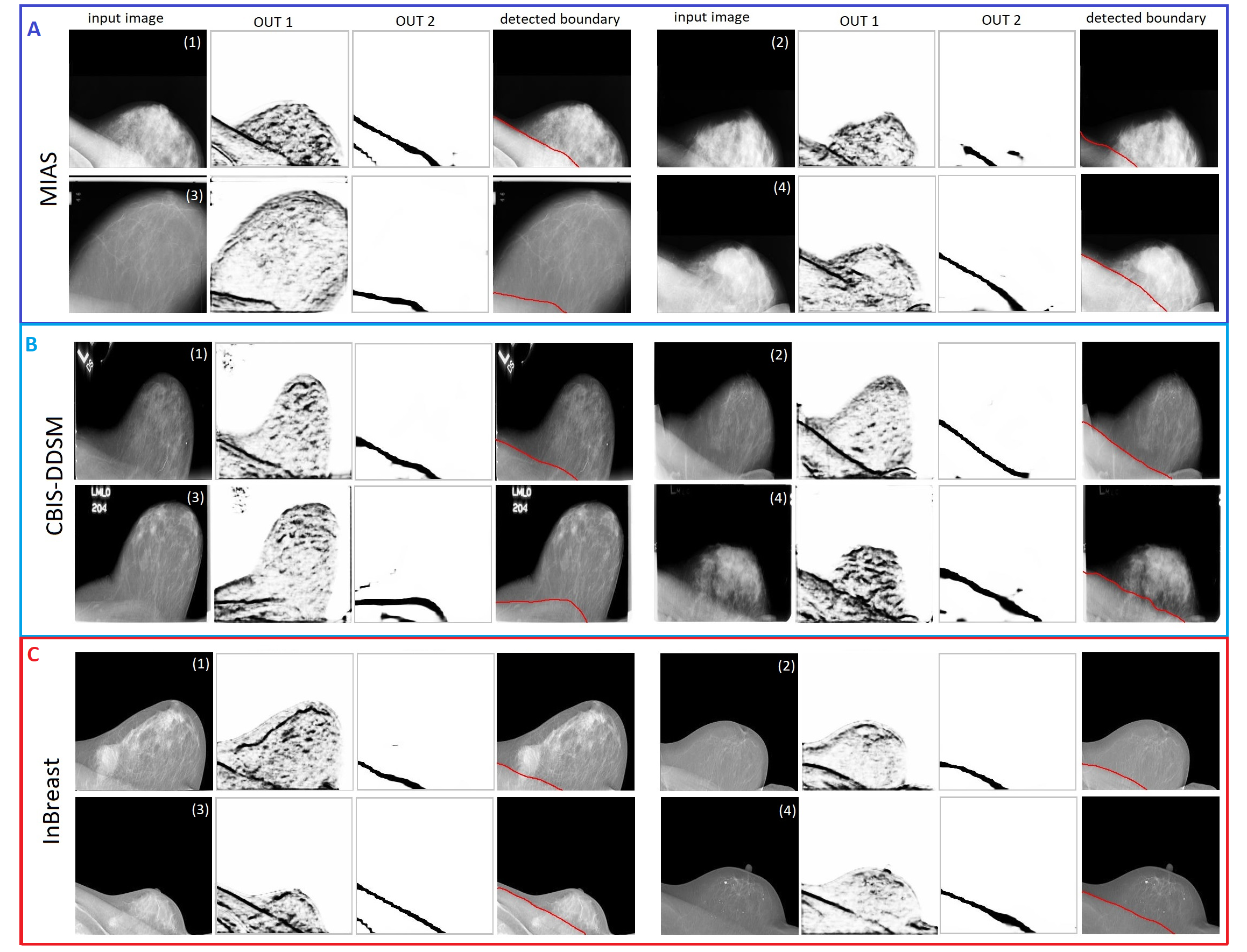}}
\caption{\textbf{Panel A}: Four examples of pectoral boundary detection showing MLO mammograms (Subplots 1-4) from the MIAS dataset along with their associated OUT1, OUT2, and detected boundary (shown in the red colour). \textbf{Panel B}: same as A, only for the CBIS-DDSM dataset; \textbf{Panel C}: same as A, only for the InBreast dataset.}
\label{fig_results_dataset}
\end{figure*}

Computation of shortest paths over weighted graphs is a well-studied problem which, in practical scenarios, is usually implemented by means of Dijkstra's algorithm or one of its many variations \cite{cormen2009introduction}. In the present work as well, we used a version of Dijkstra's algorithm which is represented by the pseudocode in Fig.~4. 

At its input, the algorithm in Fig.~4 receives the nodes $V$, binary map $B$ as well as points $S$ and $E$, while returning a set of nodes $\mathcal{P}$ which form the shortest path on $\mathcal{G}$. To simplify the notations, the pseudocode uses two ``hash functions", ${\rm dist}$ and ${\rm parent}$. For any $\pp \in V$, the functions either set or get its distance to the source node $S$ (using ${\rm dist}$) and its respective parent (using ${\rm parent}$). In particular, during initialization, the distance from $S$ to itself is set to zero (line 2), while the distances to all other nodes are set to infinity (line 3). The parental relations between the nodes, at the same time, are assume to be undefined (line 4). 

While the algorithm in Fig.~4 follows the principal structure of Dijkstra's procedure, it incorporates a few important modifications intended to maximize its numerical efficiency. In particular, the update of nodal distances to the source (lines 9-16) as well as establishing the parental relations (lines 19-25) are carried out over the 8-connected neighbourhood of each queried node $\qq$, while excluding those elements of $\mathcal{N}$ which lie outside $B$ (lines 10 and 19). As a result, the updates involves only the nodes (pixels) which are located in a close proximity of pectoralis, which substantially speeds up the computation of $\mathcal{P}$. Moreover, the updates are terminated once the end node $E$ is reached, at which point the algorithm proceeds to recovering the shortest path $\mathcal{P}$ via backtracking the parental dependencies stored by ${\rm parent}$ (lines 28-31).

Subplots A7 and B7 of Fig.~\ref{fig_results} depict the pectoral boundaries recovered by means of the proposed method for the input MLO mammograms shown in Subplots A1 and B1 of the same figure, respectively. In this case, $S$ and $E$ have been identified with the end points of the morphological skeleton of $B$, which has been observed to provide simple and stable initialization throughout all our experiments. All the principal algorithmic steps involved in the above estimation process are summarized in Fig.~5.

\section{Results}\label{sec_results}
As it was mentioned earlier, the proposed method for breast segmentation has been tested on three public datasets (MIAS \cite{suckling1994mammographic}, InBreast \cite{moreira2012inbreast} and CBIS-DDSM \cite{lee2017curated}), containing both scanned film and digital mammograms acquired from different subjects under various settings. The three panels in Fig.~\ref{fig_results_dataset} show the results of detection of the pectoral boundary, where each panel consists of four examples pertaining to the MIAS (Panel A), CBIS-DDSM (Panel B) and InBeast (Panel C) datasets. In all the cases, the binary masks $B$ were derived from OUT2 by means of standard Otsu procedure \cite{otsu1979threshold}, while the balancing parameter $\lambda$ in \eqref{balance} was set to 1.5.

In addition to input images and boundary reconstructions, Fig.~\ref{fig_results_dataset} also shows their associated OUT1 and OUT2. One can see that the largest values of OUT2 are indeed localized in a close proximity of the true pectoral boundary, thereby guaranteeing the final probability map $M$ to be free of outliers (false alarms) due to the clutter noise in OUT1. One can also see that the detection results remain consistent with respect to underlying anatomy even in the cases of low-contrast and/or incomplete observations (as depicted, e.g., in Subplots 2 \& 4 of Panel A, Subplots 1 \& 3 of Panel B or Subplots 1 \& 4 of Panel C). Moreover, the proposed method has demonstrated outstanding robustness to the presence of axillary foldings, which is another common source of artifacts impeding the process of breast segmentation (see, e.g., Subplots 1, 2, and 3 in Panels A, B, and C, respectively). 

Once a pectoral boundary is recovered, it can be combined with its related breast-air interface\footnote{Normally, this boundary is straightforward to find using, e.g., any standard edge-detection algorithm.} to produce a {\it full} breast boundary, which completes the process of breast segmentation. The interior of the resulting breast boundary can, in turn, be represented by a binary mask, which is referred below to as a breast mask. For the purpose of quantitative analysis, the quality of breast masks has been assessed in terms of a number of performance metrics. Specifically, let $R$ and $R_0$ denote an estimated breast mask and its ground truth counterpart, respectively. Also, let TP, TN, FP, and FN denote the true positive, true negative, false positive, and false negative error rates. Then, the performance metrics included:
\begin{itemize}
\item Dice similarity coefficient (DSC): $ \frac{2 |\rm R \cap \rm {R_G}|}{ |\rm R| +\rm |\rm {R_G}|}$
\item Jaccard coefficient (JAC): $ \frac{ |\rm  R \cap \rm {R_G}|}{ |\rm R \cup \rm {R_G}|}$
\item Spesificity (SPE): $ \frac{\rm {TN}}{\rm {TN+ FP}}$
\item Sensitivity (SEN): $ \frac{ \rm {TP}}{\rm {TP}+\rm {FN}}$
\item Accuracy (ACC): $ \frac{ \rm {TP + TN}}{\rm {TP+ TN + FP + FN}}$
\item False positive rate (FPR): $ \frac{ \rm {FP}}{\rm {FP+ TN}}$
\item False negative rate (FNR): $ \frac{ \rm {FN}}{\rm{ FN+ TP}}$ 
\end{itemize}

While adopting the above metrics, it is important to keep in mind some of their principal characteristics. In particular, better performance is associated with higher values of DSC, JAC, SPE, SEN and ACC and lower values of FPR and FNR. At the same time, DSC and JAC measures are known to be more sensitive in comparison to the other metrics (such as, e.g., ACC and SPE) \cite{rampun2019breast}, while JAC and DSC are known to be less sensitive to visual errors. Note that highly sensitive and highly specific algorithms rarely overlook the target they are looking for and they rarely mistake anything else for that specific target. 

\begin{table}[t]
\centering
\caption{Comparison results for three database. All metrics are presented in \% by their mean value $\pm$ one standard deviation.}
\begin{tabular}{c|ccc}
\hline
\hline
Metric  & MIAS & CBIS-DDSM & InBreast \\
\hline    
DSC & 97.59$\pm$1.73   & 97.69 $\pm$ 1.51 & 96.39 $\pm$ 2.66 \\
JAC & 95.35$\pm$3.20   & 95.52 $\pm$ 2.79 & 93.17 $\pm$ 4.79 \\
SPE & 99.83 $\pm$0.22  & 99.73 $\pm$0.34  & 99.92 $\pm$0.15  \\
SEN & 97.76$\pm$2.27   & 98.66 $\pm$ 1.74 & 94.39 $\pm$ 3.77 \\
ACC & 99.72 $\pm$0.22  & 99.51$\pm$0.35   & 99.69$\pm$0.25   \\
FPR & 0.16 $\pm$  0.22 & 0.26 $\pm$ 0.34  & 0.07 $\pm$ 0.15  \\
FNR & 2.23 $\pm$ 2.27  & 2.33 $\pm$ 1.74  & 5.60 $\pm$ 4.61 \\
\hline
\end{tabular}
\label{tab_res}
\end{table}

Table \ref{tab_res} summarizes the results of our comparative analysis by showing the mean values of the metrics (expressed in \%) plus/minus one standard deviation. One can see that, in all the cases, the values of DSC, JAC, SPE, SEN and ACC remain above 93\%, while the values of FPR and FNR are close to zero. The worst results have been observed with the InBreast dataset, with  DSC=\% 96.39, JAC=93.17\% and FNR=5.6\%. It has to be noted, however, that this dataset predominantly contains MLO mammograms with poorly defined pectoral boundaries, which are far from trivial to detect in general. At the same time, the algorithm's performance on the MIAS and CBIS-DDSM datasets has been found to be comparable (with DSC$>$97.6\%, JAC$>$95.3\%, SPE \& ACC$>$99.5\%, along with FNR$<$2.5\% and FPR$<$0.3\%).

\begin{table}[t]
\centering
\caption{Comparison results for three database by Rampun {\it et al} \cite{rampun2019breast}. All metrics are presented in \% by their mean value $\pm$ one standard deviation.}
\begin{tabular}{c|ccc}
\hline
\hline
  Metric  & MIAS         & CBIS-DDSM    & InBreast     \\
    \hline
    
DSC & 97.5$\pm$7.5   & 98.1 $\pm$ 7.1 & 95.60 $\pm$ 8.4 \\
JAC & 94.6$\pm$9.8   & 95.1 $\pm$ 9.4 & 92.6 $\pm$ 10.6 \\
SPE & 99.5 $\pm$1.2  & 99.6 $\pm$1.4  & 99.8 $\pm$1.8  \\
SEN & 98.2$\pm$7.6   & 98.3 $\pm$ 7.6 & 95.2 $\pm$ 8.6 \\
ACC & 99.3 $\pm$1.4  & 99.5$\pm$1.3   & 99.6$\pm$2.2   \\
FPR & 0.6 $\pm$  1.8 & 0.4 $\pm$ 0.6  & 0.3 $\pm$ 2.1  \\
FNR & 3.2 $\pm$ 2.9  & 3.8 $\pm$ 2.5  & 5.7 $\pm$ 6.5 \\
\hline
\end{tabular}
\label{tab_Rampun}
\end{table}

Unfortunately, direct comparison of the proposed method with available alternative approaches is problematic due to the lack of standardized ground truth segmentation. With this proviso in mind, Table \ref{tab_Rampun} summarizes the performance metrics obtained using a recent method for breast segmentation proposed by Rampun {\it et al} \cite{rampun2019breast}. Comparing these results with those in Table~\ref{tab_res}, one can see that the proposed method outperforms the reference one in most of the cases. Not less important is the fact that the proposed algorithm yields considerably smaller values of the standard deviations, which suggests that it is capable of providing more consistent estimation.

\begin{table}[t]
\centering
\caption{FPR and FNR (in \%) of reference algorithms for the MIAS dataset.}
\begin{tabular}{l |c| c c}
\hline
\hline
  Method  & Dataset & FPR & FNR \\
 \hline
Ours & MIAS(all)   & 0.16  & 2.23 \\
Rampun et al. \cite{rampun2019breast} & MIAS & 0.6  & 3.2  \\
Vikhe and Thool \cite{vikhe2017detection} & miniMIAS & 0.93  & 5.7   \\
Chen et al. \cite{chen2015shape} & miniMIAS & 1.02  & 5.63  \\
Yoon et al. \cite{yoon2016automatic} & miniMIAS & 4.51 & 5.68  \\
Bora et al. \cite{bora2016robust}& miniMIAS & 1.56  & 2.83  \\
Ferrai et al. \cite{ferrari2004identification} & miniMIAS & 0.58   & 5.77   \\
Camilus et al. \cite{camilus2010computer} & MIAS & 0.64 & 5.58  \\
\hline
\end{tabular}
\label{tab_otheralgos}
\end{table}

Additional comparative results are shown in Table~\ref{tab_otheralgos}. In particular, the table summarizes the values of FPR and FNR produced by a number of existing approaches applied to the MIAS dataset as well as to the miniMIAS dataset (which is identical to MIAS aside for the size of data images that have a dimension of $1024\times 1024$ pixels). Once again, the results indicate that the proposed method results in more accurate reconstruction in terms of FPR and FNR (which are equal to 0.16\% and 2.23\%, respectively, for the MIAS dataset). 

\begin{figure}[t]
\centering
\centerline{\includegraphics[width=8.5cm]{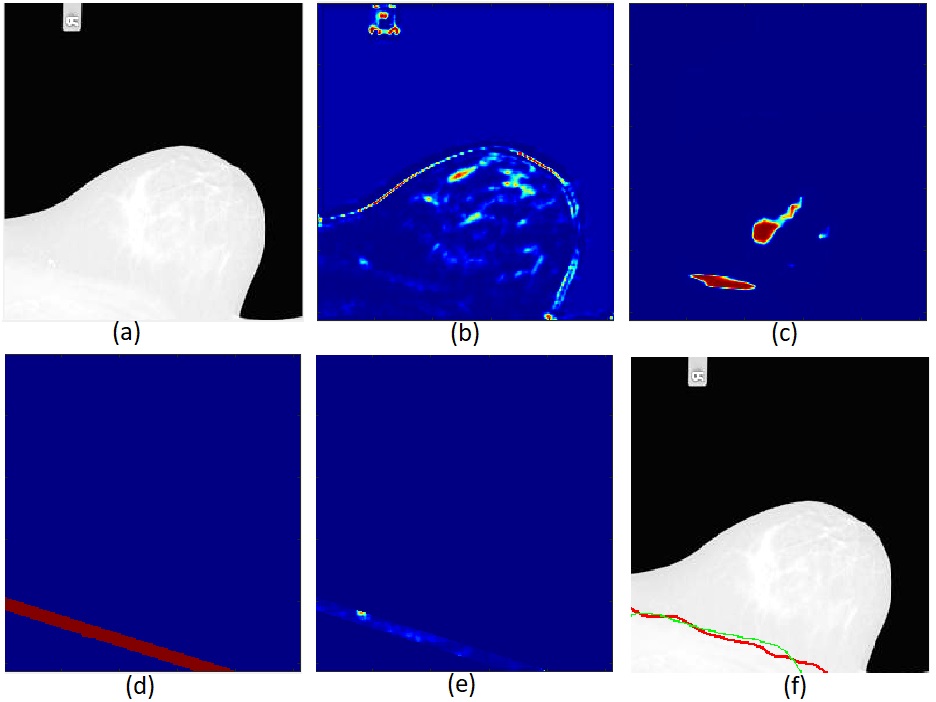}}
\caption{Detection of pectoral boundary in a low-contrast mammogram. Subplot (a): input image, Subplot (b): {\rm OUT 1}, Subplot (c): {\rm OUT 2}, Subplot (d): modified binary mask $B$, Subplot (e): final edge probability map $M$, and (f) detected boundary (red) and the ground truth (green).} 
\label{fig_extreme}
 \end{figure}

In conclusion of this section, a few words need to be added regarding the source of misclassification errors produced by the proposed algorithm. As mentioned earlier, thresholding OUT2 occasionally produces a binary mask with a fragmented longest connected component (about 10\% of all cases), as exemplified in Subplot B4 of Fig.~\ref{fig_results}. In such cases, the masks have been extended via linear extrapolation, the accuracy of which, in turn, depends on the extent of the initial mask. One of the worst-case scenarios is shown in Fig.~\ref{fig_extreme} which depicts an input mammogram (Subplot (a)) along with its related OUT1 (Subplot (b)) and OUT2 (Subplot (c)). One can see that, in this case, the longest connected component of binarized OUT2 is markedly shorter than the true pectoral boundary, being disconnected both on its left and right. It would therefore be reasonable to expect that the linearly extrapolated mask $B$ (Subplot (d)) and the final edge probability map $M$ (Subplot (e)) might provide inadequate initialization for the Dijkstra procedure in Fig.~4. However, despite the problematic data, the final result of reconstruction of the pectoral boundary shown in Subplot (f) of Fig.~\ref{fig_extreme} (red curve) appears to be in a good agreement with the ground truth (green curve), with DSC=93.85 \% and JAC=88.42\%. Thus, even though the resulting metrics are objectively lower in comparison to their mean values in Table~\ref{tab_res}, the proposed algorithm is still capable of producing meaningful and useful results.

\section{Conclusions}\label{sec_conclusion}
The use of deep neural networks for detection and analysis of salient structures in medical images has unveiled a realm of new possibilities for imaging scientists. However, properly training such networks (which are often governed by millions of tuneable parameters) is contingent on the availability of large amounts of training data. Unfortunately, such ``big data" still appear to be missing in the field of X-ray mammography. To overcome this problem, the present work advocates the use of a simplified CNN architecture (derived via ``abbreviation" of VGG16), which can be reliably optimized based on relatively small training sets. In view of its limited description power (as a result of using a substantially smaller number of network parameters), the proposed CNN is not expected to produce the final result, but rather to provide reliable initialization for the second part of post-processing that takes advantage of Dijkstras's algorithm to complete the task.    

The proposed two-step reconstruction of the breast mask is advantageous in a number of ways. First, the use of CNN eliminates the need for analytical assumptions regarding the geometry of pectoral boundary, which are prone to errors due to the effects of inter-subject variability. Consequently, the CNN-based processing remains stable and consistent across a variety of different pectoral anatomies, imaging contrasts, and image acquisition methods (e.g., scanned film vs digital mammography). Furthermore, the subsequent use of graph-based processing allows one to recover the pectoral boundary at a single-pixel resolution in a fully automatic way as well.

Our experimental results have demonstrated a superior performance of the proposed solution over state-of-the-art, especially when it comes to difficult cases of low-contrast mammograms with partially occluded or barely distinguishable pectoral boundaries. The quantitative assessments of the solution performed on three different datasets (i.e., MIAS, CBIS-DDSM, and InBreast) resulted in the average values of DSC, FPR and FNR equal to 97.22 $\pm$ 1.96\% , 0.16 $\pm$ 0.23\% and 3.38 $\pm$ 2.87\%, respectively, which suggests that the proposed methodology constitutes a viable alternative to existing approaches to the problem of breast segmentation.


\bibliographystyle{IEEEbib}
\bibliography{main}

\end{document}